# Identify the cells' nuclei based on the deep learning neural network

Tianyang Zhang, Rui Ma

University of Massachusetts Amherst, Department of Electrical and Computer Engineering
tianyangzhan@umass.edu, ruima@umass.edu

## Abstract

*Identify the cells' nuclei is the important point for most medical analyses. To assist doctors finding the accurate cell' nuclei location automatically is highly demanded in the clinical practice. Recently, fully convolutional neural network (FCNs) serve as the back-bone in many image segmentation, like liver and tumer segmentation [10] in medical field, human body block in technical filed. The cells' nuclei identification task is also kind of image segmentation. To achieve this, we prefer to use deep learning algorithms. we construct three general frameworks, one is Mask Region-based Convolutional Neural Network (Mask RCNN) [2], which has the high performance in many image segmentations, one is U-net [1], which has the high generalization performance on small dataset and the other is DenseUNet [10], which is mixture network architecture with Dense Net [9] and U-net. we compare the performance of these three frameworks. And we evaluated our method on the dataset of data science bowl 2018 challenge [14]. For single model without any ensemble, they all have good performance.*

*Index Terms—Mask RCNN, DenseUNet, U-net, deep learning, hybrid features, image augmentation, semantic segmentation*

1. Introduction

Identifying the cells' nuclei is the starting point for most analyses because most of the human body's 30 trillion cells contain a nucleus full of DNA, the genetic code that programs each cell. Identifying nuclei allow researchers to identify each individual cell in a sample, and by measuring how cells react to various treatments, the researcher can understand the underlying biological processes at work.
Identifying nuclei is the typical classification and detection problem. In the last two years, deep convolutional network has outperformed state of art in many visual recognition tasks. While convolutional network existed for a long time, the success of convolutional network was limited if the size of the available training set is not enough or the size of considered network is limited. Also, the image segmentation still a big challenge due to the pixel-wise segmentation. For the cells' nuclei identification problem, we all meet these two challenges, one is that the size of our training dataset is limited; the training dataset contain only 640 microscope images, each image includes variety type of cells. The other is that the cells' nuclei identification task is kind of image semantic segmentation, we should segment the background and cells' nuclei at the same time. We trained three kind of network on this task, one is U-net [1], this network is good for handling small dataset task and achieve good performance on very different biomedical segmentation applications. Olaf et al trained this model just given 30 images to win the cell tracking challenge 2015. One is Mask Region-based Convolutional Neural Network (Mask RCNN) [2], this network is very powerful in many instance segmentations. We modified this model and set the classifier 2, only background and cells' nuclei, to fit our condition. The other one is DenseUNet [10], this network using Dense Net [9] to extract the feature, and use the similar U-net upsampling architecture to upsampling the feature map. Xiaomeng Li et al use this model to achieve perfect performance for liver segmentation in medical field. We do some modification based on these three advanced models, and evaluated these three model on data science bowl 2018 challenge. They all have a good performance.

2. Problem statement

The goal of this task is to segment each cells' nuclei and background. The dataset we use contains 640 microscope images, see graph 1. The images were acquired under a variety of conditions and vary in the cell type, magnification, and imaging modality. Each image is represented by an associated imageId, and there are two folders in dataset, one folder contains the original image and one folder contains the segmented mask of each nucleus, and each mask contains one nucleus. We split the dataset into training dataset and evaluation dataset. The evaluation dataset is the 20 percent of the dataset by random selection and the left is our training dataset.

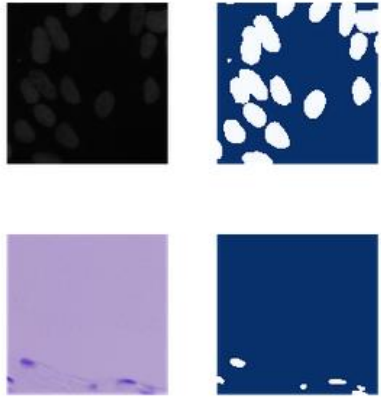

Fig. 1 example of the microscope images left and the segmented masks of the cells' nucleus after we merge all separated nucleus mask in one image right

3. Image augmentation

Data augmentation is essential to teach the network the desired invariance and robustness properties, especially for small dataset. 640 images are not enough for us to train a very deep neural network, It will easily cause the overfitting problem. We tried a lot of augmentation methods to enlarge our dataset. As a result, we choose several methods to implement. The graph 2 displayed the images after augmentation and corresponding masks.

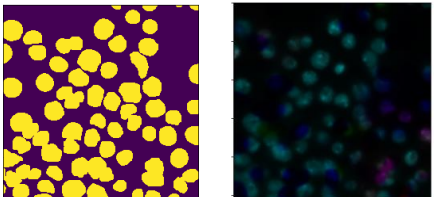

Fig. 2 example of the microscope images with augmentations right and the corresponding segmented masks left

### 3.1.1  Blurring

Blurring is an important aspect of augmentation since it trains our model to abstract useful features (cores' position) even if images are not very clear, in other words, it improves the robustness of our model. In practice, we tried motion blurring (very blur) and median blurring into the image. We set 0.1 probability to motion blurring and 0.3 to median blurring.

### 3.1.2  Graying

Changing colored images into gray images helps improving recognition accuracy and reducing computation time in processing convolution layers.

### 3.1.3  Embossing

Image embossing is a technique in which each pixel of an image is replaced either by a highlight or a shadow, depending on light/dark boundaries on the original image. Low contrast areas are replaced by a gray background. The filtered image will represent the rate of color change at each location of the original image. Implementing this method helps to locate the cell zones in the images.

### 3.1.4  Channels Rearranging

changing the channels' order is helpful to improve color resistance of our model. In practice, we change the order RGB into GBR or BGR with a probability of 0.3.

### 3.1.5  Sharpening, contrast and brightness

Sharpening, increasing contrast and brightness are all methods to address the key features. This allows our model to learn more images that contains certain key features.

### 3.1.6  Zoom in/out and rotation

We also set random zoom in/out and rotation method to produce more training data. We set the zoom ration 0.1 and rotation with 90/180/270.

4. Related work

U-net: the traditional U-net architecture consists of a contracting path and an expansive path. We prefer you can look at Olaf et al paper [1] and they do excellent work for that. The contracting path is the typical architecture of a convolutional network; the expansive path consists the upsampling of the feature map. For each upsampling step, the upper layer feature map will concatenate with the correspondingly feature map from the contracting path.

R-CNN: The Region-based CNN (R-CNN) approach [3] to bounding-box object detection is to attend to a manageable number of candidate object region [2, 4, 5] and evaluated convolutional network independently on each RoI (Regions of Interest). It allows RoIs on feature map using RoIPool. Faster R-CNN [6] advanced this stream with a Region Proposal Network(RPN) by learning attention mechanism. Mask RCNN implement the RoIAlign operation for extracting a small feature map from each RoI. Instead of using any quantization of the RoI boundaries or bins, RoIAlign use binear interpolation to calculate the exact values of the four regularly sampled location in each bin [2].

Dense Net: the traditional Dense Net [9] have 4 dense blocks, and 3 transition layers. Each dense block with different amount of conv blocks, and each conv block contains two group of 1x1 conv followed by 3x3 conv. Each transition layer has one 1x1 conv followed by 2x2 average pool. We use the Dense net 121 as our backbone for DenseUNet, the number of conv blocks in each dense block is 6, 12, 24, 16. Dense Net is a perfect framework for feature extraction, it has highway path for each feature map in each dense block.

DenseUNet: the DenseUNet using Dense Net for feature extraction, and Unet-like expansive path for feature upsamling. The dense connection between layers is employed within each dense block to ensure the information flow and UNet-like lone ranges connection links from the encoding part (feature extraction part) and the decoding part (feature expansive path) to preserve low-level information.

Evaluation matrix [12]: the evaluation matrix for cells' nuclei identification was used mean average precision at different intersection over union (IoU) thresholds. The IoU of a proposed set of object pixels and a set of true object pixels is calculated as:

$$IoU(A,B) = \frac{A \cap B}{A \cup B}$$

The metric sweeps over a range of IoU thresholds, at each point calculating an average precision value, the threshold values range from 0.5 to 0.95 with a step size of 0.05. In other words, at a threshold of 0.5, a predicted object is considered a "hit" if its intersection over union with a ground truth object is greater than 0.5.

At each threshold value tt, a precision value is calculated based on the number of true positives (TP), false negatives (FN), and false positives (FP) resulting from comparing the predicted object to all ground truth objects:

$$\frac{TP(t)}{TP(t) + FP(t) + FN(t)}$$

A true positive is counted when a single predicted object matches a ground truth object with an IoU above the threshold. A false positive indicates a predicted object had no associated ground truth object. A false negative indicates a ground truth object had no associated predicted object. The average precision of a single image is then calculated as the mean of the above precision values at each IoU threshold:

$$\frac{1}{|thresholds|} \sum_t \frac{TP(t)}{TP(t) + FP(t) + FN(t)}$$

5. Network Architecture

U-net: we modified our framework from traditional U-net. The table 1 show our networks architecture. Our network still consists of contracting path and expansive path; the contracting path consist of 10 convolutional layers and 4 max pooling layers (graph 3). We repeated two 3x3 convolution layers with exponential linear unit (ELU) four times, the stride step is 1 and each time the number of kernels are 8, 16, 32, 64. The 2x2 max pooling operation with stride 2 was added after each two repeated convolution layers. The last two layers of contracting path are two conv and the number of kernels are 128. The expansive path consists of six 3x3 convolutional layers, three 2x2 deconvolution layers [11], three dropout layers and three concatenate layers, we repeated one 2x2 deconvolution layer followed with one concatenate layer, one dropout layer and two 3x3 convolution layers three times, each concatenate layer will concatenation with related feature map in the contracting path. The final layer is a 1x1 convolution layer with sigmoid activation function used to map 32 feature vectors after expensive path to the desired numbers of classes. In total, our U-net has 20 layers. The input image of U-net is 512x512x3, and the output mask of U-net is 128x128x3. Our U-net is unbalanced between our inputs and outputs, that's for reducing upsampling information loss and memory saving.

Mask RCNN: Thanks to the martterport [13] et al, they do the Mask RCNN framework and we modified the architecture from them framework. Mask RCNN has the multiple architecture [2], (1) the backbone layer used for extract the feature from entire image, we use the ResNet [7] network with 50 layers to extract feature, also, we added the Feature Pyramid Network (FPN) [8] after ResNet-50, FPN used a top-down lateral connection to build an in-network feature pyramid from a single-scale input, the ResNet plus FPN backbone would extract RoI in different feature pyramid, it will give us more accuracy for feature extraction. (2) The network head architecture; the mask RCNN add the mask branch which is parallel with bounding box and classification branch [2], this branch has four 3x3 convs with stride 1, one 2x2 deconvs with stride 2, and the output layer is 1x1 convs. We modified the head architecture of Mask RCNN and keep the backbone layer using ResNet-50 and FPN. To support training multiple images per batch we resize all images to same size (512x512x3), and we set the 2 classes, nucleus and background. We use momentum as our optimizer, the learning rate set 0.001 and momentum set 0.9. we also use the mini mask which resize the mask to smaller size (56x56x1) for memory saving.

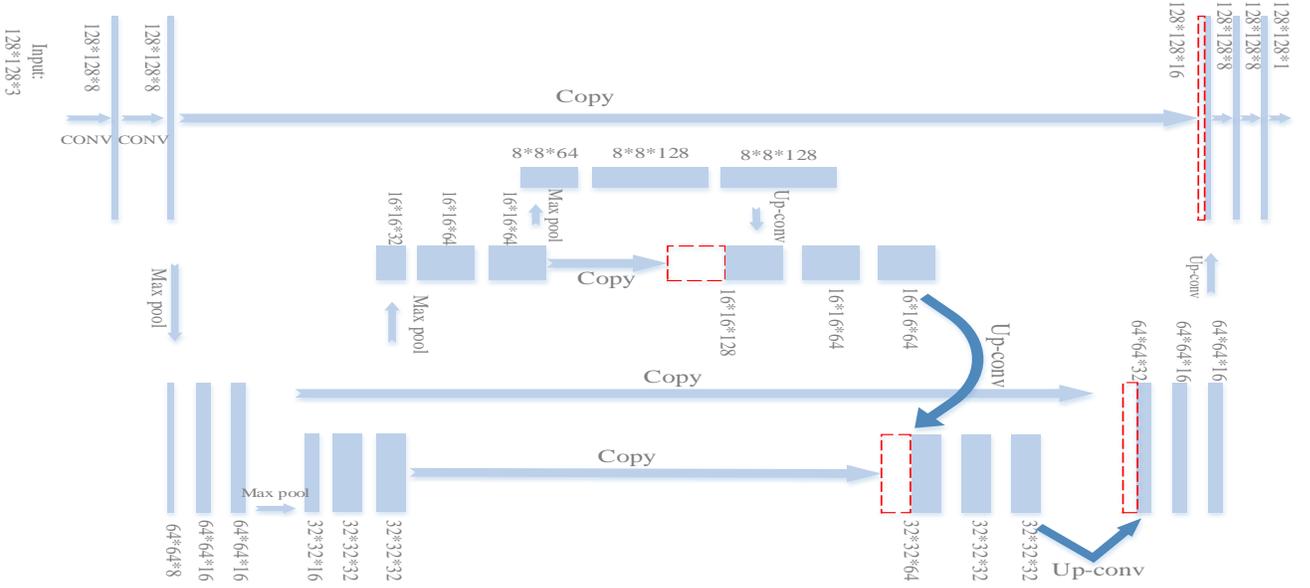

Fig. 3 U-Net Architecture

Table 1 Architecture of the proposed U-Net, symbol [ ] denotes the concatnate layer

| Layers | Feature Size | U-Net |
|---|---|---|
| Input | 512×512 | - |
| Convolution (1) | 512×512 | 3x3 conv x2 |
| Pooling | 256×256 | 2x2 max pooling |
| Convolution (2) | 256×256 | 3x3 conv x2 |
| Pooling | 128×128 | 2x2 max pooling |
| Convolution (3) | 128×128 | 3x3 conv x2 |
| Pooling | 64×64 | 2x2 max pooling |
| Convolution (4) | 64×64 | 3x3 conv x2 |
| Pooling | 32×32 | 2x2 max pooling |
| Convolution (5) | 32×32 | 3x3 conv x2 |
| Umsampling Layer (1) | 64x64 | 2x2 Deconv, [conv layer (4)], dropout, 3x3 conv x2 |
| Umsampling Layer (2) | 128×128 | 2x2 Deconv, [conv layer (3)], dropout, 3x3 conv x2 |
| Convolution (6) | 128x128 | 1x1 conv |

DenseUNet: we also make some changes from traditional DenseUNet. The exact network show in the table 2. Instead of using 4 dense block and 3 transition layers in DenseNet-121, we only use first three dense block with two transition layers. The dense block 3 followed by the expansive path, the expansive path consists of one upsampling layer, one concatenate layer, one batch normalization layer, one activation layer, one dropout layer and one convolution layer. The concatenate layer would concatenation with the last layers of each dense block, for example, the concatenate layer 1 would concatenation with the dense block 2, and concatenate layer 2 would concatenation with the dense block 1. Activation layer we use exponential linear unit (ELU), experiment demonstrate that ELU have better performance than ReLU on DenseUNet. We repeat this construction two times and the last layer followed by one 1x1 convolution layer with sigmoid activation function to output the desired number of classes. The gragh 4 illustrate the framework of the DenseUNet.

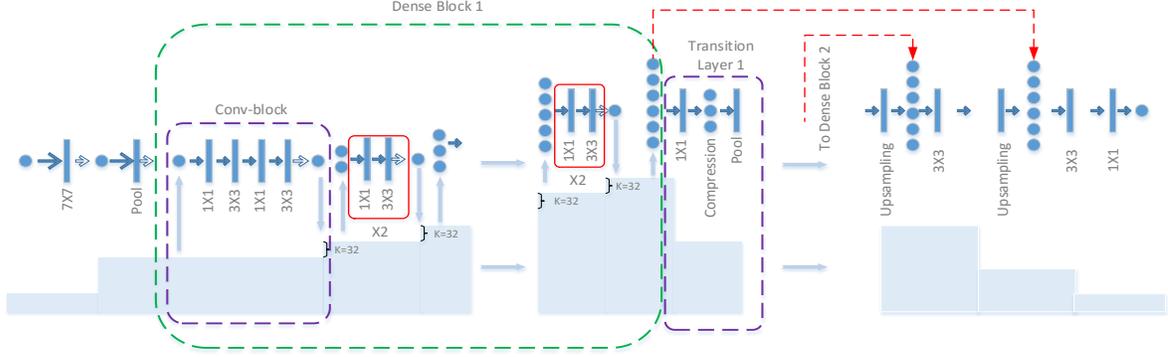

Fig. 4 The illustration of DenseUNet

Table 2 Architecture of the proposed DenseUNet, the symbol [ ] denotes the long range UNet summation connections. The { } denotes the conv block. The ( ) denotes one group of 1x1 conv followed by 3x3 conv. Note that each 'convs' corresponds the sequence BN-ReLU-Conv and 'conv2' represent sequence BN- ELU-Dropout-Conv

| Layers | Feature Size | DenseUNet |
|---|---|---|
| input | 512×512 | - |
| Convolution (1) | 256×256 | 7×7 convs, stride 2 |
| Pooling | 128×128 | 3×3 max pooling, stride 2 |
| Dense Block (1) | 128×128 | conv block {(1x1 convs, 3x3 convs) x 2} x6 |
| Transition Layer (1) | 128×128 | 1×1 convs |
|  | 64×64 | 2×2 average pool, stride 2 |
| Dense Block (2) | 64×64 | conv block {(1x1 convs, 3x3 convs) x 2} x 12 |
| Transition Layer (2) | 64×64 | 1×1 convs |
|  | 32×32 | 2×2 average pool, stride 2 |
| Dense Block (3) | 32×32 | conv block {(1x1 convs, 3x3 convs) x 2} x 16 |
| Upsampling Layer (1) | 64×64 | 2×2 Upsampling [Dense Block 2], 96 conv2 |
| Upsampling Layer (2) | 128×128 | 2×2 Upsampling [Dense Block 1], 64 conv2 |
| Convolution (2) | 128×128 | 1×1 conv |

6. Experiments

We demonstrate the application of our three frameworks to two different datasets, one is the original dataset without any image augmentation, the dataset contains 640 images and the images were acquired under a variety of conditions and vary in the cell type, magnification, and imaging modality. Other one is the dataset we have done the images augmentation; we copy the same images 5 times, and each same image we do different kind of augmentations randomly, all images transfer to gray, and this dataset we have total 3200 images. The dataset was provided by the data science bowl 2018 challenge.

We split our dataset into two part, one is training dataset which is 80 percent of dataset by random selection and other is validation dataset. On the original image dataset, we only test the U-net and DenseUNet frameworks, we set the dropout rate 0.5, the higher rate to prevent overfitting happen much earlier, and we use the early stop mechanism, the patience of the loss without any improve on the validation dataset is 2. After doing that, the mAP rate of U-net on the test dataset is 0.311, the DenseUNet is 0.381. On the images augmentation dataset, we trained the Mask RCNN, U-net and DenseUNet, the Mask RCNN get 0.476 mIoU rate, DenseUNet get 0.443 and U-net get 0.325. We provide the full results in table 3.

Table 3 Segmentation Results on data science bowl 2018 challenge

| | Input Size | Output Size | mean average precision (mAP) | mean average precision (mAP) |
|---|---|---|---|---|
| U-Net | 512×512×3 | 128×128×1 | 0.311 | 0.325 |
| Mask R-CNN | 512×512×3 | 56×56×1 | - | 0.476 |
| DenseUNet | 512×512×3 | 128×128×1 | 0.381 | 0.442 |

7. Conclusion

We present three frameworks, DenseUNet, UNet and Mask RCNN for data science challenge 2018, the DenseUNet and Mask RCNN architecture achieves good performance on this challenge, we use NVidia tesla p100 (16GB) to train the model. We tried different kind of images augmentations and modified the model in different ways, that give us a lot of experiences. In addition, these three kind of architecture is inherently general and can be easily extended to other segmentation applications.